\documentclass[10pt]{article}
\usepackage[utf8]{inputenc}
\usepackage{times}
\usepackage{newtxmath}
\usepackage{tgcursor}
\usepackage[margin=1in]{geometry}
 
\usepackage{mathtools}
\usepackage{xcolor}
\usepackage{hyperref}
\definecolor{bluegray}{rgb}{0.3, 0.5, 0.7}
\hypersetup{
    colorlinks=true,
    linkcolor=bluegray,
    urlcolor=bluegray,
    citecolor=bluegray
}
\usepackage{cleveref}
\usepackage{subfig}
\usepackage{booktabs}
\usepackage{tikz}
\usepackage{natbib}

\setcitestyle{authoryear,open={(},close={)}}
\usepackage[T1]{fontenc}
\usepackage{inconsolata}

\usepackage{amsmath}
\usepackage{graphicx}
\usepackage{subfig}
\usepackage{placeins}
\usepackage{floatrow}
\definecolor{probable}{HTML}{7782b0}
\definecolor{improbable}{HTML}{86b7ca}
\definecolor{impossiblephysics}{HTML}{dca12b}
\definecolor{impossiblemagic}{HTML}{e7be6d}
\definecolor{inconceivablesemantic}{HTML}{d9596d}
\definecolor{inconceivablesyntactic}{HTML}{eba9a5}
\usepackage{linguex}
\newcommand{\ngram}{$n$-gram}
\newcommand{\ngrams}{$n$-grams}
\newcommand{\infinigram}{$\infty$-gram}
\newcommand{\gpt}{GPT-2}
\newcommand{\llama}{Llama-3}
\newcommand{\olmo}{OLMo}

\usepackage{mathtools}
\DeclareMathOperator*{\argmax}{arg\,max}
\newcommand{\defeq}{\vcentcolon=}

\begin{document}

\vspace*{1cm}

\begin{center}
    \Large{Shades of Zero: Distinguishing Impossibility from Inconceivability} 
    \bigskip
    \bigskip
    
    \large \today
    
    \bigskip
    
    \large{Jennifer Hu\textsuperscript{1}, Felix Sosa\textsuperscript{2}, Tomer Ullman\textsuperscript{1,2}}

    \bigskip
    
    \normalsize
    \textsuperscript{1}Kempner Institute for the Study of Natural and Artificial Intelligence, Harvard University \\
    \textsuperscript{2}Department of Psychology, Harvard University
\end{center}

\vspace{3cm}

\noindent \textbf{Author contributions}

\begin{table}[ht]
    \centering
    \small
    \begin{tabular}{cccccc} \toprule
        Author & Designed research & Performed research & Analyzed data & Wrote the paper \\ \midrule
        JH & \checkmark & \checkmark & \checkmark & \checkmark \\
        FS & \checkmark & \checkmark & \checkmark & \checkmark \\
        TU & \checkmark & & \checkmark & \checkmark \\ \bottomrule
    \end{tabular}
\end{table}

\vspace{2cm}

\noindent \textbf{Corresponding authors}\\
Jennifer Hu (jenniferhu@fas.harvard.edu)

\vspace{2cm}

\noindent \textbf{Data availability statement}\\
Our code, experimental stimuli, and data are publicly available at \url{https://github.com/jennhu/shades-of-zero}. 

\thispagestyle{empty}

\section*{Abstract}
Some things are impossible, but some things may be even more impossible than impossible. Levitating a feather using one's mind is impossible in our world, but fits into our intuitive theories of possible worlds, whereas levitating a feather using the number five cannot be conceived in any possible world (``inconceivable''). While prior work has examined the distinction between improbable and impossible events, there has been little empirical research on inconceivability. Here, we investigate whether people maintain a distinction between impossibility and inconceivability, and how such distinctions might be made. We find that people can readily distinguish the impossible from the inconceivable, using categorization studies similar to those used to investigate the differences between impossible and improbable (Experiment 1). However, this distinction is not explained by people's subjective ratings of event likelihood, which are near zero and indistinguishable between impossible and inconceivable event descriptions (Experiment 2). Finally, we ask whether the probabilities assigned to event descriptions by statistical language models (LMs) can be used to separate modal categories, and whether these probabilities align with people's ratings (Experiment 3). We find high-level similarities between people and LMs: both distinguish among impossible and inconceivable event descriptions, and LM-derived string probabilities predict people's ratings of event likelihood across modal categories. Our findings suggest that fine-grained knowledge about exceedingly rare events (i.e., the impossible and inconceivable) may be learned via statistical learning over linguistic forms, yet leave open the question of whether people represent the distinction between impossible and inconceivable as a difference not of \emph{degree}, but of \emph{kind}. \\

\noindent\textbf{Keywords:} 
impossibility; inconceivability; modal reasoning; language models; event knowledge

\section*{Highlights}
\begin{itemize}
    \item People readily distinguish between impossible and inconceivable events
    \item Language model probabilities also distinguish impossible and inconceivable
    \item Language model probabilities broadly align with people's ratings of event likelihood
\end{itemize}

\newpage

\section{Introduction} \label{sec:intro}

Some things are impossible. You cannot levitate a feather with your mind, no matter how hard you try. And yet, some things are more impossible than others. Levitating a feather with one's mind is in some sense easier than levitating a rock \citep{shtulman_explanatory_2017, mccoy_judgments_2019}. Such graded judgments of impossibility are the topic of ongoing study in cognitive science, philosophy, and cognitive development. The idea is that people's understanding of imaginary worlds is rooted in their understanding of the real world \citep{byrne2007rational}, so studying what makes things easier or harder in the imagination can reveal people's understanding of everyday reality. As part of this research, there have been different explanations for what makes some things seem more impossible than others. The explanations are not mutually exclusive, and include (for example) moves across ontological hierarchies \citep{griffiths_revealing_2015}, causal violations \citep{shtulman_explanatory_2017}, perceived similarity \citep{goulding2023perceived}, and violations of core knowledge and intuitive physics \citep{mccoy_judgments_2019, lewry2021intuitions}. 

But, just as there is a dividing line between the merely improbable and the truly impossible, there may be a category of events even more impossible than impossible. Levitating a feather with one's mind is impossible in our world, but can still be imagined as occurring in a fictional world, and fits into our intuitive theories of possible worlds \citep{lewis1986plurality}. In contrast, events like ``levitating a feather using the number five'' or ``finding the square-root of a dog'' cannot be evaluated, construed, or imagined in any possible world.\footnote{Some readers may object that they can imagine various transformations of the example sentences to make sense of them, such as supposing ``the number five'' refers to a particular blow-drier in a set. We refer to such transformations as \textit{coercions}, and return to them in the Discussion. \label{fn:coercions}}
Borrowing from philosophy, we refer to such events as \textit{inconceivable} \citep{gendler_conceivability_2002}. Our suggestion is that just as it has been fruitful for cognitive science to study people's understanding of the impossible, it is useful to study people's understanding of the inconceivable.  

While the relationship between conceivability and possibility has been the topic of much philosophical research \citep[e.g.,][]{chalmers_conscious_1996,balog_conceivability_1999,gendler_conceivability_2002}, there has been less empirical and computational cognitive science study of inconceivability. The most closely related line of work has investigated children's distinction between events that are \emph{impossible} (e.g., walking on water) and events that are possible but highly \emph{improbable} (e.g., growing a beard down to one's toes, or singing Jingle Bells at a birthday party) \citep[e.g.,][]{komatsu_childrens_1986,browne_preschoolers_2004,shtulman_improbable_2007,shtulman_development_2009,lane_childrens_2016,goulding_similarity_2021,goulding_development_2023}. A main finding across these studies is that while even preschool-age children can distinguish impossible and probable events, they tend to judge improbable (or expectation-violating) events as \emph{impossible}, and that the ability to distinguish the improbable and impossible is refined with age. This has been taken by some to suggest that young children try to conceive of an improbable event, fail, and judge their inability to conceive of it as evidence that the event cannot occur at all \citep{harris2021early}. However, such a distinction does not explain the impossible/inconceivable divide for adults: adults are easily able to imagine things they judge to be impossible, at least in our world. These findings in general leave open the question of whether adults naturally treat impossible and inconceivable as distinct modal categories (as they do for impossible and improbable), or whether people naturally treat inconceivability as simply an instance of impossibility.

We begin by investigating whether people can readily distinguish the impossible from the inconceivable, using categorization studies similar to those used to investigate the differences between improbable and impossible. These categorization tests follow the logic of the cognitive development studies mentioned above, which conclude that young children do not distinguish improbable from impossible events early in development, but refine their understanding over early childhood \citep[e.g.,][]{shtulman_improbable_2007,shtulman_development_2009,shtulman_differentiating_2018}.
In our first experiment (\Cref{sec:exp1}), we introduce a novel set of stimuli featuring event descriptions in four different modal categories (probable, improbable, impossible, and inconceivable), and validate them by examining whether people intuitively and reliably distinguish inconceivable event descriptions from event descriptions belonging to the other modal categories. We find that people are highly consistent in their categorizations, suggesting the set of inconceivable event descriptions is easily distinguished from the sets of impossible and improbable event descriptions.

Having established that people readily do make this distinction, we then examine how this distinction might be made. Broadly, there are two functional accounts for how the distinction between impossibility and inconceivability may work, corresponding to a distinction in kind, and a distinction of degree. A functional framework in support of the distinction in kind would argue that encountering the inconceivable  is similar to the mind encountering a category error \citep{magidor_category_2009, ryle_concept_1949}, leading to either termination of further processing, or the need to coerce the meaning of the original event. Category errors may in turn be similar to ``type errors'' encountered by type-based computer programs \citep{magidor_category_2009, sosa_type_2022}.  
This general view also connects to a literature on type-theoretic distinctions in semantic theory \citep[e.g.,][]{montague_proper_1973,luo_formal_2012,bekki_representing_2014,chatzikyriakidis_formal_2020,sutton_types_2024}, and psycholinguistic studies of selectional restrictions \citep{chomsky_aspects_1965,katz_structure_1963} in language processing \citep[e.g.,][]{warren_investigating_2007,warren_comprehending_2015,paczynski_multiple_2012,sitnikova_two_2008}. An alternative possibility is that people have a single, graded notion of probability, where all modal events exist on a spectrum, including the improbable, impossible, and inconceivable. Each event is processed and assigned some probability of occurring, and modal categories are then read out by defining a straightforward transformation on top of the underlying probabilities. As a simplified example, one could define thresholds on probability values such that improbable events are one-in-a-hundred, impossible are one-in-a-million, and inconceivable are one-in-a-trillion. This view also connects to prior proposals that selectional restrictions can be defined based on statistical associations \citep{resnik_semantic_1993,resnik_selectional_1996}, and evidence that world knowledge violations elicit similar patterns of neural activity as selectional restriction violations \citep[e.g.,][]{hagoort_integration_2004,matsuki_event-based_2011}. Our experiments investigated two potential versions of the second account (``difference in degree''), but we do not resolve the debate of whether inconceivable differs from impossible in kind or degree. We return to the different mechanisms for distinguishing inconceivable and impossible in the Discussion (\Cref{sec:discussion}).

In our second experiment (\Cref{sec:exp2}), we ask people for subjective ratings of the likelihood of the events from Experiment 1, on a continuous scale from 0 to 100. We find that people overwhelmingly rate impossible and inconceivable events at the bottom of the scale, and the ratings across these categories cannot be distinguished statistically. This suggests that people's modal distinctions between impossible and inconceivable are based on information beyond simple perceptions (or transformations) of event likelihood, which weighs against a basic version of the ``graded'' view -- namely, that modal categories are read off based on thresholds upon probability.

We then investigate another version of the graded view by asking to what extent distinctions between modal categories can be made based on the statistical patterns of language.
From a cognitive perspective, one proposal has been that people try to imagine or simulate an event occurring, and use the difficulty of imagination to judge the possibility of the event \citep{kahneman1981simulation,gendler_conceivability_2002, harris2021early}. 
It is possible to argue that event descriptions that we frequently encounter are easier to simulate or imagine, and event descriptions that are very infrequently encountered will be harder to simulate. 
To investigate this view, we examine whether statistical language models (LMs) display similar behavioral signatures as humans regarding the distinction between inconceivable event descriptions and other modal categories. LMs are trained with the objective of predicting sequences of tokens, which they learn to do by observing vast amounts of text typically obtained from Internet posts, news datasets, and books. Through this paradigm, LMs may implicitly learn the latent properties of the world that make certain linguistic expressions more or less likely. For example, since people communicate about events \citep{mcrae_people_2009} and physical observations \citep{louwerse_symbol_2011,louwerse_knowing_2018}, it may be reasonable to expect that language itself contains structured information about the world. Indeed, prior work has shown that LMs may capture important aspects of commonsense and world knowledge \citep[see][for review]{chang_language_2023}, including the distinction between possible and impossible events \citep{kauf_event_2023}, and the structure of perceptual spaces \citep{abdou_can_2021,patel_mapping_2022,merullo_linearly_2023}. 
The autoregressive next-token-prediction objective used to train LMs also has connections to human behaviors during language comprehension, which involve prediction about upcoming linguistic content \citep[e.g.,][]{altmann_incremental_1999,levy_expectation-based_2008,kutas_thirty_2011,smith_effect_2013,shain_large-scale_2023} and the integration of generalized event knowledge \citep[e.g.,][]{mcrae_people_2009,bicknell_effects_2010,matsuki_event-based_2011}. It is therefore argued that LMs may learn structured information about the world (and events occurring in the world) in service of optimizing the objective of next-word prediction. 

From a machine learning perspective, the question of whether the distinction between impossibility and inconceivability can be learned based on statistical patterns is also highly of interest. If both impossible and inconceivable events occur with vanishingly small frequency, it is not obvious how LMs would learn to distinguish them in string probability space. 
Indeed, our work's focus on distinctions within the region of low probabilities -- ``shades of zero'' -- differs from two previous major approaches to evaluating LMs. The first prior approach focuses on distinguishing between broad categories which are expected to be associated with high or low probabilities -- for example, by performing targeted comparisons between sentences that are grammatical/ungrammatical \citep[e.g.,][]{marvin_targeted_2018,warstadt_blimp_2020,hu_systematic_2020}, or describe possible/impossible events \citep[e.g.,][]{kauf_event_2023}. The second approach involves studying the behavior of LMs on naturally occurring sentences, which are expected to fall within the region of the string distribution with most probability mass -- for example, by examining the alignment between model-derived string distributions and human reading times on naturalistic sentences \citep[e.g.,][]{smith_effect_2013,brothers_word_2021,shain_large-scale_2023,hofmann_language_2022}. While these approaches are reasonable and useful, we suggest that studying distinctions within the near-zero region of the probability distribution -- such as the distinction between sentences describing impossible and inconceivable events -- can provide new insights into the patterns and commonsense knowledge learned by LMs, and how their representations of event likelihood relate to people's conceptual categories. 

In our third experiment (\Cref{sec:exp3}), we test whether LMs separate the modal categories from our stimuli through the probabilities assigned to string event descriptions, and whether these probabilities align with humans' subjective ratings of event likelihood (as we measured in Experiment 2). We evaluate five Transformer-based language models from the GPT-2 \citep{radford_language_2019} and Llama 3 \citep{aimeta_llama_2024} model families, ranging in size from 124 million to 70 billion parameters, as well as a strong non-neural baseline \citep{liu_infini-gram_2024}. We find that modal categories are strongly separated within the event description probabilities estimated by the neural models, but not within the probabilities estimated by the non-neural baseline. Similarly, the subjective event likelihood ratings provided by humans in Experiment 2 are predicted by the neural models' probabilities (but not the baseline model's probabilities) -- although models' probabilities vary widely across impossible and inconceivable event descriptions, while people assign these extremely low likelihood ratings. We additionally test how distinctions between modal categories emerge over the course of training for a single OLMo model \citep{groeneveld_olmo_2024}, and find that the distinction between inconceivable and impossible emerges before the distinction between impossible and improbable. These analyses contribute a new empirical investigation of commonsense and event knowledge in modern LMs, which is a central topic in artificial intelligence \citep{zhou_evaluating_2020,li-etal-2022-systematic,chang_language_2023}. In particular, we demonstrate that LMs capture a distinction within the distribution of non-possible events (impossible versus inconceivable), which goes beyond previous studies focusing on distinctions either within the distribution of \emph{possible} events (likely versus unlikely) or \emph{across} the boundary of possible and impossible \citep[e.g.,][]{wang_modeling_2018,kauf_event_2023,kauf_comparing_2024}.

Overall, our findings reveal high-level similarities between the way that humans and statistical language models treat inconceivability. First, both humans and models show an ability to categorize probable, improbable, impossible, and inconceivable event descriptions (comparing Experiments 1 \& 3). Second, the probabilities assigned to event descriptions by neural LMs predict humans' subjective ratings of event likelihood (comparing Experiments 2 \& 3). And third, the distinction between improbable and impossible emerges later during LM training than the distinctions between other pairs of categories, which is qualitatively consistent with findings in developmental psychology \citep[e.g.,][]{shtulman_improbable_2007}. However, our findings also raise questions about \emph{how} humans make distinctions between impossible and inconceivable when their subjective likelihoods are indistinguishable (all near zero), and suggest future directions for investigating children's perceptions of inconceivability. We return to these issues in the Discussion (\Cref{sec:discussion}), once we have the results in hand.

\section{Experiment 1: Modal classification task} \label{sec:exp1}

We first developed a novel set of materials, which included short English-language event descriptions across four modal categories: probable, improbable, impossible, or inconceivable. Our main questions of interest are whether humans categorize events in a way that is consistent with the underlying coding of conditions, and what kind of error patterns humans make. This experiment also serves to validate the construction of our stimuli, which we use in later experiments. 

\paragraph{Data availability statement} All of our code, materials, and data are publicly available at \url{https://github.com/jennhu/shades-of-zero}.

\subsection{Stimuli}

\begin{table*}[t]
    \centering
    \small
    \begin{tabular}{llllll} \toprule
        {\bf\color{black}{Prefix}} & {\bf\color{probable}{Probable}} & {\bf\color{improbable}{Improbable}} & {\bf\color{impossiblephysics}{Impossible}} & {\bf\color{inconceivablesemantic}{Inconceivable}} & {\bf\color{inconceivablesyntactic}{Inconceivable-Syntax}} \\ \midrule
        baking a cake inside & an oven & an airfryer & a freezer & a sigh & grasp \\
        chilling a drink using & ice & snow & fire & yesterday & onto \\
        washing your hair with & shampoo & detergent & a rainbow & applause & if \\
        smashing a pumpkin using & a hammer & a boulder & a leaf & a number & remember \\ \bottomrule
    \end{tabular}
    \caption{Sample items used in our experiments. Each item (row) consists of a shared prefix and a set of continuations across five conditions. The Inconceivable-Syntax condition was never used to evaluate humans, and only used in Experiment 3 as a baseline for the language model evaluation.}
    \label{tab:sample-items}
\end{table*}

We manually constructed a set of 70 items designed to cover several modal categories (see examples in \Cref{tab:sample-items}). Each item consists of a shared prefix denoting a commonplace event with a transitive verb and object, as well as the beginning of a phrase describing how the event occurs (e.g., ``baking a cake inside''). No explicit subject is specified. Each item prefix is associated with five candidate continuations, each of which completes the phrase describing how the event occurs. These continuations reflect different types of modal relationships to the prefix. Since the region that modulates the modal category of the event occurs at the end of the phrase, these stimuli are well-suited to test autoregressive (i.e., left-to-right) language models, which condition on preceding context to predict the subsequent tokens. While prior cognitive work on modal distinctions has primarily focused on probable, improbable, and impossible events, we additionally include inconceivable events.  
We describe each condition in greater detail below. 

Continuations in the \textbf{Probable} and \textbf{Improbable} conditions reflected events that are possible given the physical laws of the real world. We refer to these as the ``possible'' conditions. Probable continuations were chosen to be prototypical or highly expected (e.g., ``baking a cake inside an \emph{oven}'', or ``washing your hair with \emph{shampoo}''). Improbable continuations were intuitively unconventional, but did not violate any physical constraints (e.g., ``baking a cake inside an \emph{airfryer}'', or ``washing your hair with \emph{detergent}''). 

In the \textbf{Impossible} condition, continuations make the event impossible because of physical constraints in the real world. For example, ``baking cake inside a \emph{freezer}'' violates thermodynamics. This condition is most similar to what prior studies have called ``violations of world knowledge'' (cf.~violations of \emph{word} knowledge, or selectional restrictions). But, some instances do not neatly fit into this framework. For example, ``washing your hair with \emph{a rainbow}'' could be a violation of world knowledge, or a violation of the restrictions on the verb ``to wash'' (i.e., $[+\textsc{Liquid}]$). 

Finally, we considered \textbf{Inconceivable} continuations, which take the form of an abstract concept or non-physical noun when the event described in the prefix expects a concrete noun (e.g., ``baking a cake inside \emph{a sigh}'' or ``chilling a drink using \emph{yesterday}''). This condition is the most similar to selectional restriction violations in the linguistics literature \citep{chomsky_aspects_1965,katz_structure_1963}. 

\subsection{Methods}

We recruited $N=149$ participants on Prolific, who were based in the United States and native speakers of English by self-report. Participants were compensated at a rate corresponding to \$12/hour. Each participant saw each of the 70 items, one item per trial. On each trial, participants saw a prefix and a continuation in one of the four conditions (e.g., ``Baking a cake using an oven''). Participants were asked to categorize the stimulus into one of four categories (by pressing keyboard buttons): probable, improbable, impossible, or nonsense. We used the label ``nonsense'' instead of ``inconceivable'' to avoid jargon, while capturing a similar intuition. After responding on each trial, participants saw a screen saying ``Your response has been logged'' for 1 second. Each participant saw a nearly equal number of trials in each condition (i.e., 17 or 18 trials for each of the 4 main conditions). The order of trials and conditions was randomized. Prior to the experiment, participants were familiarized with definitions and examples of each category, and were required to correctly complete 8 practice trials before beginning the critical trials.

\subsection{Results}

\begin{figure}[t]
    \centering
    \subfloat[\label{fig:human-results-accuracy}]{
        \includegraphics[width=0.6\linewidth]{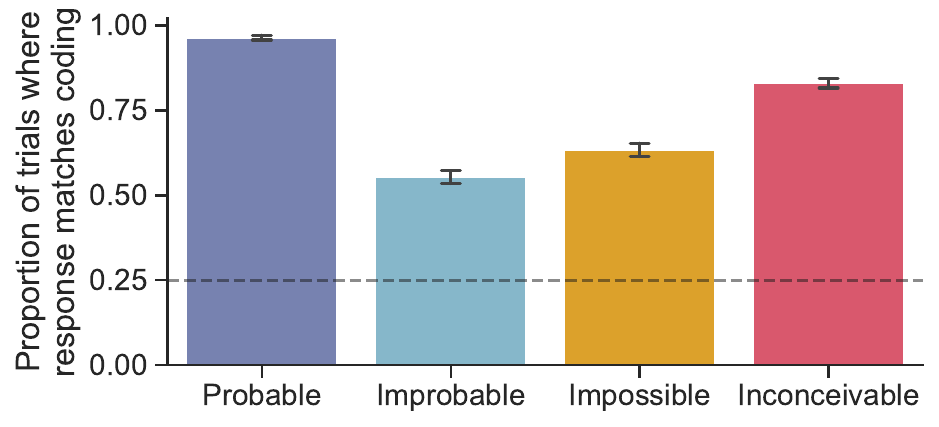}
    }%
    \subfloat[\label{fig:human-results-confusion}]{
        \includegraphics[width=0.38\linewidth]{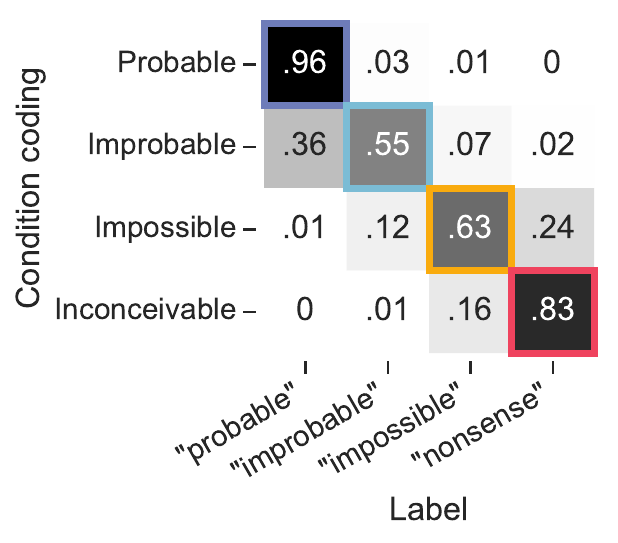}
    }
    \caption{Classification results from Experiment 1. (A) Proportion of trials where human responses matched the underlying condition coding. Dashed line indicates chance performance (25\%). Error bars indicate bootstrapped 95\% CI. (B) Distribution of responses within each condition. Highlighted cells indicate proportion of trials where responses match the condition coding (as shown in (A)).}
\end{figure}

First, we ask whether humans categorize the events in a way that is consistent with the researcher-defined coding of the materials. \Cref{fig:human-results-accuracy} shows the proportion of trials where people's labels matched the underlying condition coding. In the Probable condition, participants agreed with our coding in nearly all cases (96\%). Participants matched our coding for the overwhelming majority of cases in the Inconceivable condition (83\%), and in the majority of cases for the Improbable (55\%) and Impossible (63\%) conditions.
These results suggest that people's judgments align with our stimulus coding at a broad level, validating the construction of our materials. Furthermore, this provides evidence that people can intuitively categorize inconceivable items, mirroring and extending the findings of prior cognitive studies showing that people distinguish the impossible from the improbable \citep[e.g.,][]{shtulman_improbable_2007}.

Next, we ask what kind of choices people make, when they categorize events in a way that is inconsistent with our researcher-defined labels. \Cref{fig:human-results-confusion} shows the distribution of responses in each condition, averaged across trials. The errors (i.e., trials where the participant's label of an event does not match our coding) are not random, but reflect structured patterns of confusability between neighboring modal categories. For example, when participants did not label the Improbable items as improbable, they were most likely to label them as probable (36\%). This is reasonable, as the distinction between probable and improbable is a matter of degree, and there is inherent imprecision of the term improbable. In addition, subjective estimates of the probabilities of possible (i.e., probable or improbable) events may vary based on each individual's experiences and environments. 

We also found evidence for confusability between modal categories which, \emph{a priori}, might be expected to have firm boundaries. For example, Improbable items were labeled as impossible 7\% of the time, and Impossible items were labeled as improbable 12\% of the time. While the majority of responses label Impossible items as impossible and Inconceivable items as nonsense, we also observe some confusability between impossibility and inconceivability. 24\% of responses labeled Impossible as nonsense, and 16\% labeled Inconceivable as impossible. These patterns suggest that, while people do readily distinguish inconceivable from impossible (and other modal categories), the boundaries between these categories may be graded. In the next experiment, we more directly test this idea by investigating whether people assign graded ratings of likelihood to impossible and inconceivable events.

\section{Experiment 2: Subjective ratings of event likelihood} \label{sec:exp2}

In our second experiment, we asked human participants to rate how likely events are to occur, using a slider scale.
This task tests a basic version of the view that the distinction between impossibility and inconceivability is a difference in \emph{degree}. If people's modal distinctions between impossible and inconceivable are based on simple thresholds on top of subjective event likelihood, then we would expect to find separation between the likelihood ratings assigned to impossible and inconceivable events. If there is no separation in these ratings, then this would suggest that people's distinctions are made on another basis.

\subsection{Methods}

We used the same set of stimuli as in Experiment 1. We recruited $N=50$ US-based participants on Prolific, with a self-reported native language of English. Participants were compensated at an hourly rate of \$12. Each participant saw each of the 70 items, one item per trial. On each trial, participants saw a question of the form ``How likely is it for someone to [EVENT DESCRIPTION]?'', where the event description follows one of the four conditions (e.g., ``How likely is it for someone to bake a cake inside a sigh?''). Their task was to respond using a slider with endpoints marked ``Extremely unlikely'' and ``Extremely likely'', which were internally coded as 0 and 100. No increments or numbers were shown on the slider. Each participant saw a roughly equal number of trials in each condition (i.e., either 17 or 18 trials for each of the 4 conditions). 

We normalized participants' responses within items, to control for people's perceived prior probabilities of the prefixes, independent of the continuations. For example, ``baking a cake'' might be more likely \emph{a priori} than ``docking a boat'', and so it could be the case that someone would rate the Probable continuation of ``docking a boat'' as less likely than the Improbable continuation of ``baking a cake''.

\begin{figure}[t]
    \centering
    \subfloat[\label{fig:ratings-raw}]{
        \includegraphics[height=1.5in]{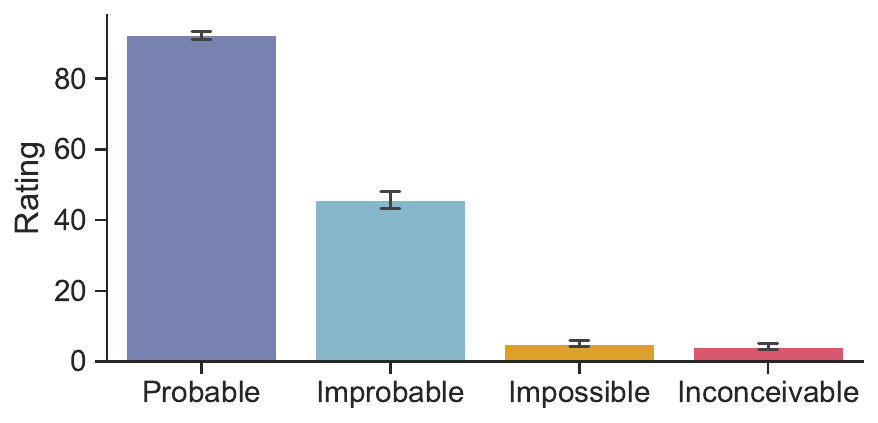}
    }%
    \subfloat[\label{fig:ratings-normalized}]{
        \includegraphics[height=1.5in]{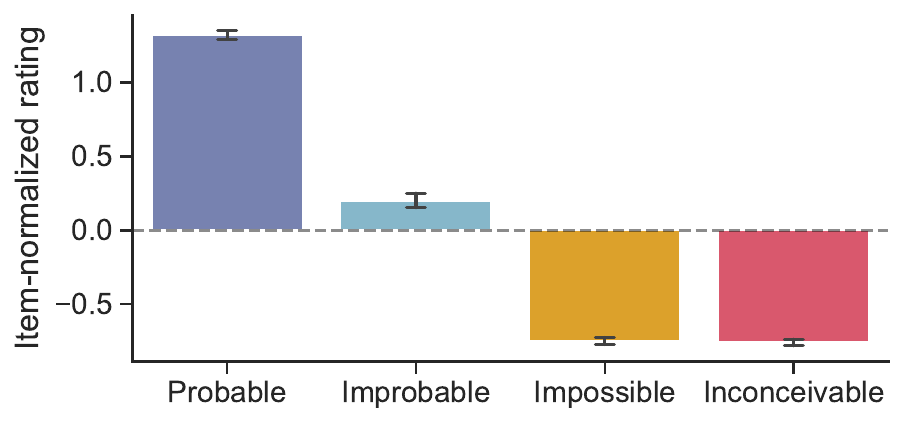}
    }
    \caption{Subjective ratings of event likelihood from Experiment 2. (A) Raw and (B) normalized (within-item) ratings, averaged over items in each condition. Error bars denote bootstrapped 95\% CIs.}
    \label{fig:ratings}
\end{figure}

\subsection{Results}

\Cref{fig:ratings-raw,fig:ratings-normalized} show raw and item-normalized ratings across conditions, respectively. We first turn to the raw ratings (on a 0-100 scale). The Probable event descriptions received near-ceiling likelihood ratings (mean rating: 92.2), the Improbable items were near the middle (mean rating: 45.6), and the Impossible and Inconceivable items were near the floor of the slider scale (Impossible mean rating: 5.0; Inconceivable mean rating: 4.2). We find a similar pattern for the within-item normalized ratings as well (\Cref{fig:ratings-normalized}). Importantly, the Impossible and Inconceivable event descriptions did not differ in their rating distributions (item-normalized ratings; two-sample Kolmogorov-Smirnov test: $D=0.04, p=0.37$; Mann-Whitney U test: $U=379911.5, p=0.79$).

To summarize our results thus far, we have found that people can reliably distinguish impossible and inconceivable events (Experiment 1), but they assign them indistinguishable near-zero likelihood of occurring (Experiment 2). This suggests that people's categorizations are not based purely on differences in perceived likelihood, which rules out a basic ``threshold-probability'' version of the view that the distinction between impossibility and inconceivability is one of \emph{degree}. However, these results do not rule out the graded account entirely, nor do they provide direct support for the view that this distinction is a difference in \emph{kind} (analogous, for example, to a type error). We test an alternate version of the graded account in the next experiment, by investigating the treatment of model categories in string probabilities estimated by statistical language models. We return to the open question of how people make the distinction between impossible and inconceivable in the Discussion (\Cref{sec:discussion}).

\section{Experiment 3: Language model evaluation} \label{sec:exp3}

Experiment 1 established that people can intuitively distinguish between events in the categories of probable, improbable, impossible, and inconceivable. Experiment 2 showed that people assign essentially equal (near-zero) subjective likelihood ratings to sentences describing impossible and inconceivable events. In our third  experiment, we investigate whether statistical language models (LMs) treat these modal distinctions in a similar way to humans: do the string-level probabilities of event descriptions separate modal categories, even impossible and inconceivable? And do these probabilities align with humans' subjective ratings of event likelihood? 

\subsection{Methods} 

\subsubsection{Stimuli}

We use the same stimuli as in Experiments 1 and 2. However, we add an additional condition as a baseline: Inconceivable-Syntax. This condition features a continuation that makes the full phrase ungrammatical, typically by featuring a verb, adverb, or preposition instead of the expected noun (e.g., ``chilling a drink using \emph{at}''). We emphasize that this condition is not the focus of our investigation of impossibility versus inconceivability -- instead, it serves as a baseline, since we expect language models (LMs) to recognize these expressions as ill-formed.\footnote{Whether category errors constitute a syntactic error is a long-standing debate that we do not resolve here; see \citet{magidor_category_2013}.} In the context of the current experiment, this means that we expect LMs to reliably assign lower probabilities to these continuations, as would be consistent with a large body of work on targeted syntactic evaluation of LMs \citep[e.g.,][]{marvin_targeted_2018,hu_systematic_2020,warstadt_blimp_2020}. Examples of the Inconceivable-Syntax condition are shown in \Cref{tab:sample-items}. Again, this condition is only used for the model evaluation, and not the human experiments.

\subsubsection{Models}

\begin{table}[t]
    \centering
    \footnotesize
    \subfloat[]{
        \begin{tabular}{lllrl} \toprule
            Model name & Model family & Model identifier & \# parameters & Training data size \\ \midrule
            \infinigram & \ngram{} \citep{liu_infini-gram_2024} & v4\_dolma-v1\_7\_llama & -- & 2.6 T tokens \\
            \gpt{} & \gpt{} \citep{radford_language_2019} & gpt2 & 124 M & 40 GB text \\
            \gpt{} Med & \gpt{} \citep{radford_language_2019} & gpt2-medium & 355 M & 40 GB text \\
            \gpt{} XL & \gpt{} \citep{radford_language_2019} & gpt2-xl & 1.5 B & 40 GB text \\
            \llama{} 8B & \llama{} \citep{aimeta_llama_2024} & meta-llama/Meta-Llama-3-8B & 8 B & 15 T tokens \\
            \llama{} 70B & \llama{} \citep{aimeta_llama_2024} & meta-llama/Meta-Llama-3-70B & 70 B & 15 T tokens \\ \bottomrule
        \end{tabular}
    }\quad%
    \subfloat[]{
        \begin{tabular}{rlrrr} \toprule
            Checkpoint & Model identifier & \# parameters & \# steps & \# training tokens \\ \midrule
            1 & allenai/OLMo-7B-0424-hf & 7B & 500 & 2B \\
            2 & allenai/OLMo-7B-0424-hf & 7B & 1000 & 4B \\
            3 & allenai/OLMo-7B-0424-hf & 7B & 2000 & 8B \\
            4 & allenai/OLMo-7B-0424-hf & 7B & 4000 & 16B \\
            5 & allenai/OLMo-7B-0424-hf & 7B & 8500 & 35B \\
            6 & allenai/OLMo-7B-0424-hf & 7B & 17500 & 73B \\
            7 & allenai/OLMo-7B-0424-hf & 7B & 36000 & 150B \\
            8 & allenai/OLMo-7B-0424-hf & 7B & 72500 & 303B \\ 
            9 & allenai/OLMo-7B-0424-hf & 7B & 151500 & 635B \\ 
            10 & allenai/OLMo-7B-0424-hf & 7B & 477000 & 2T \\ \bottomrule
        \end{tabular}
    }
    \caption{Details of language models evaluated in Experiment 3. (A) Models tested in size manipulation experiment. Model identifiers correspond to Huggingface pretrained model paths for the neural models, and the API index for \infinigram. (B) Checkpoints of \olmo{} tested in training time experiment.}
    \label{tab:models}
\end{table}

We manipulate two properties that might affect models' ability to differentiate between the categories of interest: size (number of parameters), and training time. 

For our investigation of model size, we evaluated six open-source language models: five neural network-based language models, and one \ngram{} baseline. We include the \ngram{} baseline to capture a statistical LM based purely on co-occurrence counts in a large corpus. If this model captures human-like behaviors in distinguishing the modal categories, this suggests that these distinctions could be explained by different frequencies of direct experience with these events. If not, this would suggest that the neural network-based LMs are doing something more sophisticated than tracking frequency counts or token co-occurrences to assign probabilities to event descriptions.

Our five neural models include three size variants of the \gpt{} family \citep{radford_language_2019} and two size variants of the \llama{} family \citep{aimeta_llama_2024}. Collectively, the models range in size from 124 million to 70 billion parameters, spanning multiple orders of magnitude. Each neural model is an autoregressive language model based on the Transformer architecture \citep{vaswani_attention_2017}, and publicly available through the Huggingface Transformers library \citep{wolf_transformers_2020}. 

As a strong non-neural baseline, we evaluate the infini-gram model \citep[``\infinigram'';][]{liu_infini-gram_2024}, which estimates arbitrarily large \ngram{} counts based on the Dolma-v1.7 corpus \citep{soldaini_dolma_2024}. This corpus contains 2.6 trillion tokens created by the Llama-2 tokenizer \citep{touvron_llama_2023}, substantially larger than other corpora commonly used to estimate \ngrams, such as COCA \citep{davies_corpus_2008} or Google Books \citep{juola_google_2022}. 

For our investigation of training time, we evaluated 10 geometrically-spaced training checkpoints of \olmo-7B \citep{groeneveld_olmo_2024}, ranging from 500 steps (corresponding to 2 billion training tokens) to 477,000 steps (corresponding to 2 trillion tokens). \olmo{} was also trained on Dolma-v1.7, the same corpus used to estimate \ngram{} counts and train the \infinigram{} model.

\subsubsection{Estimating surprisal}

We use the same stimuli as in Experiment 1 (see \Cref{tab:sample-items} for examples). To estimate how (un)expected an event is, we measure the surprisal $S$ \citep{hale_probabilistic_2001,levy_expectation-based_2008}, or negative log probability, of the continuation $c$ conditioned on its prefix $p$:
\begin{equation}
    S(c | p) \defeq -\log P(c | p)
\end{equation}
If a model has learned to represent event probabilities in a way that conforms to the normative coding of our stimuli, then it should assign lowest surprisal to Probable continuations, higher surprisal to Improbable continuations, and highest surprisal to Impossible and Inconceivable continuations.

On a technical note, we are interested in probabilities assigned to words or multi-word expressions, while models compute probabilities at the level of \emph{tokens}. To obtain the log probability of a continuation, we take the mean log probability over all tokens within the continuation.\footnote{We appended a period at the end of each continuation when evaluating the neural LMs, in order to signal the end of a sentence. This is done because in certain cases the event description could be continued in a way that changes the modality of the event (e.g., ``washing your hair with applause'' could be continued with ``playing on a speaker in the bathroom''). We did not include the final period when evaluating the \infinigram{} model or estimating \ngram{} frequencies in order to avoid sparsity issues.} This controls for the potential confound of length, as continuations that are split into more tokens will likely have lower probabilities overall. 

To deal with sparsity issues for \ngram{} estimates, \infinigram{} performs a version of backoff where the model conditions on the longest suffix in the prefix that appears in the training data. Since this backoff method still leaves open the possibility of zero counts, we also use additive smoothing to deal with sparsity. That is, given corpus $\mathcal{D}$ and vocabulary $V$, the \infinigram{} probability of the $i$th token $t_i$ in a sequence is given by
\begin{equation}
    P_{\infty-\text{gram}}\left(t_i | t_1, \dots, t_{i-1}\right) \defeq \frac{\text{count}\left(t_{i-(n-1)}, \dots, t_{i-1}, t_i | \mathcal{D}\right) + k}{\text{count}\left(t_{i-(n-1)}, \dots, t_{i-1} | \mathcal{D}\right) + k|V|},
\end{equation}
where
\begin{equation}
    n \defeq \argmax_{n' \in [1, i]} \left\{\text{count}\left(t_{i-(n'-1)}, \dots, t_{i-1} | \mathcal{D}\right) > 0 \right\}.
\end{equation}
In practice, we use $k=0.1$ as a compromise between full pseudocounts ($k=1$) and no smoothing ($k=0$), taking into account the large vocabulary size of the tokenizer (under which larger $k$ might distort the probability estimates more). 

Finally, by comparing the surprisal of different continuations conditioned on the same prefix, there is the potential confound of observing differences driven by frequency effects. For example, we might observe
\begin{equation}
    S(\text{an airfryer } | \text{ Baking a cake using}) > S(\text{an oven } | \text{ Baking a cake using}),
\end{equation}
simply because ``an airfryer'' (the Improbable continuation) occurs less frequently than ``an oven'' (the Probable continuation) in text corpora. One way to address this would be to counterbalance the items, such that each continuation is seen in every condition, which has been a standard practice in other LM evaluations. However, our manipulation of inconceivability relies on pairing concrete-action prefixes with abstract continuations, meaning that the inconceivable continuations cannot be paired with any prefix that would make the described event conceivable. 
Therefore, we instead validate our materials by comparing the \ngram{} frequency counts of each continuation across conditions, and test for whether the condition significantly affects the surprisal values after controlling for \ngram{} frequencies. We estimated the \ngram{} frequency counts of the continuations using Dolma-v1.7 (2.6 trillion tokens), the same corpus used to train the \infinigram{} model. 

\subsection{Results}

\begin{figure}[t]
    \centering
    \includegraphics[width=\linewidth]{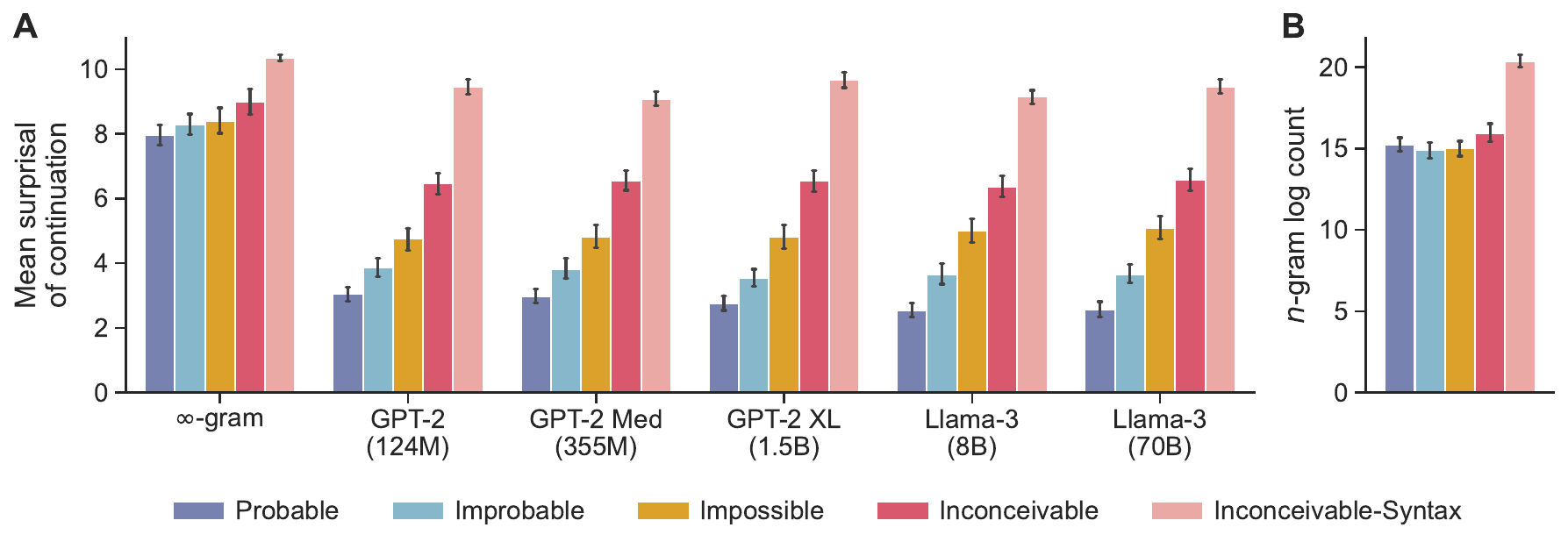}
     \caption{(A) Mean surprisal values (i.e., negative log probability averaged over tokens) assigned by our tested language models to continuations in each condition. (B) Log $n$-gram count of continuations in each condition, estimated based on Llama-2 tokenization of Dolma-v1.7 \citep{liu_infini-gram_2024}. Error bars denote bootstrapped 95\% CIs.} 
     \label{fig:surprisal-ngram}
\end{figure}

\subsubsection{Distinguishing modal categories}

First, we compare LMs' probability distributions to people's categorization behaviors in Experiment 1: do LM surprisals distinguish between the modal categories of probable, improbable, impossible, and inconceivable?

\paragraph{Size manipulation.} \Cref{fig:surprisal-ngram}A shows the surprisal values for continuations in each condition, averaged across items. Numerically, for each model, the condition-level mean surprisals are ordered in the following way (mirroring the ordering of the x-axis): Probable $<$ Improbable $<$ Impossible $<$ Inconceivable $<$ Inconceivable-Syntax. We additionally analyzed pairwise comparisons of surprisals across conditions \emph{within} each item (\Cref{sec:appendix-exp3-item-level}, \Cref{fig:item-level-surprisal-heatmap}), and the item-level surprisal comparisons also largely follow the condition ordering discussed above, except for the \infinigram{} model. Turning to the frequency counts of each continuation, \Cref{fig:surprisal-ngram}B shows the mean log count of occurrences of each \ngram{} across conditions. The counts are in fact higher in the Inconceivable and Inconceivable-Syntax conditions than the other conditions, which should bias surprisal values to be \emph{lower} in these conditions, since LMs are sensitive to frequency effects \citep{mccoy_embers_2024}. If we find higher surprisals in the Inconceivable conditions relative to the others, this could not be explained purely by frequencies, suggesting that LMs are representing meaningful distinctions across the modal categories.

For each tested language model, we fit a linear mixed-effects regression model using the \texttt{lme4} package in R to test the effect of condition and \ngram{} frequency counts. We additionally included a random intercept for each item, as specified by the following formula:
\begin{equation}
    \texttt{meanSurprisal $\sim$ condition + logNgramCount + (1 | item)} \label{eq:lmer_condition_ngram}
\end{equation}
The condition variable was backward-difference coded, such that the mean surprisal for one condition is compared to the mean surprisal of the prior adjacent condition. In other words, the variable has four levels, corresponding to the following four pairwise comparisons: (1) Probable $<$ Improbable; (2) Improbable $<$ Impossible; (3) Impossible $<$ Inconceivable; and (4) Inconceivable $<$ Inconceivable-Syntax. For each neural model, we find a strong positive effect of each these pairwise condition comparisons. For the baseline \infinigram{} model, we find a small but significant effect for the Inconceivable $<$ Inconceivable-Syntax comparison, but no effect for the other three condition comparisons. Full results from each regression model are shown in \Cref{sec:appendix-exp3-stats}, \Cref{tab:lmer}.

Overall, these results suggest that the modal categories from our materials (probable, improbable, impossible, inconceivable) can be distinguished from each other in string probability space. In addition, the success of the neural models in contrast to the failure of the \infinigram{} -- the strongest available \ngram-based baseline -- suggests that this ability is not simply based on word co-occurrences.

\begin{figure}[t]
    \centering
    \includegraphics[width=0.7\linewidth]{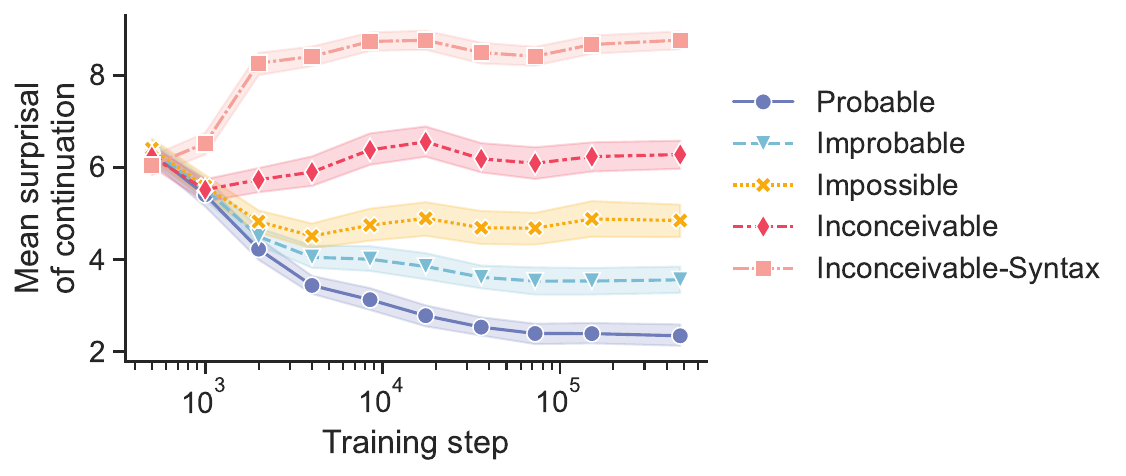}
     \caption{Mean surprisal values assigned to continuations in each condition by 10 intermediate checkpoints of OLMo-7B. Error bands indicate bootstrapped 95\% CIs.} 
     \label{fig:surprisal-training}
\end{figure}

\paragraph{Training time manipulation.}

Next, we analyze the separability of modal categories over training time of a single language model: \olmo-7B. \Cref{fig:surprisal-training} shows the mean surprisal assigned to continuations in each condition across training checkpoints. First, we note that the syntactic violations (Inconceivable-Syntax) separate the earliest during training, which is perhaps unsurprising. Interestingly, the separation between Improbable/Impossible seems to occur later during training than the separation between Impossible/Inconceivable and Probable/Improbable. The relationship between Probable, Improbable, and Impossible seems broadly consistent with patterns of findings in the developmental literature \citep{shtulman_improbable_2007,shtulman_development_2009}, where younger children struggle to distinguish improbable and impossible events.
Our findings also suggest a novel prediction that has been largely unexplored in child psychology, and could be tested in future experiments: the distinction between impossible and inconceivable events would occur earlier in development than the distinction between improbable and impossible.

\begin{figure}[t]
    \centering
    \includegraphics[width=\linewidth]{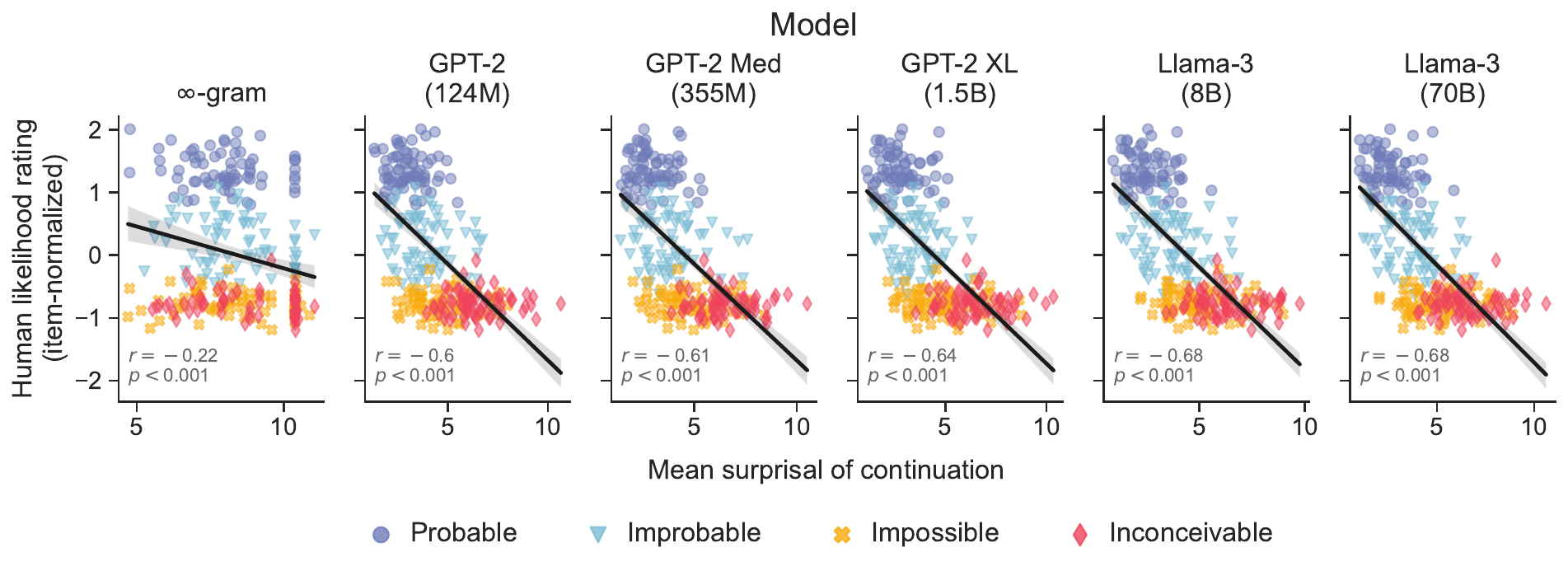}
    \caption{Human ratings (y-axis) versus surprisal assigned by language models to continuations in each condition (x-axis). Subplots are annotated with Pearson $r$ correlation coefficients. While the linear correlation captures the overall trend, it does not capture the relevant distinction between impossible (yellow x's) and inconceivable (red diamonds).}
    \label{fig:rating-vs-surprisal}
\end{figure}

\subsubsection{Surprisal versus human ratings}

Next, we compared the surprisal values derived by language models and the human ratings measured in Experiment 2 (\Cref{sec:exp2}). To analyze the predictive power of model surprisals, we fit the following linear mixed-effects regression model for each language model: 
\begin{equation}
    \texttt{rating $\sim$ surprisal + logNgramCount + (1 | item) + (1 | subject)} 
    \label{eq:lmer_rating_surprisal}
\end{equation}
Overall, the model surprisals strongly predict human ratings, even when taking frequencies into account (\infinigram: $t=-13.83, p<2e-16$; \gpt: $t=-48.193, p<2e-16$; \gpt{} Med: $t=-49.668, p<2e-16$; \gpt{} XL: $t=-52.450, p<2e-16$; \llama{} 8B: $t=-58.953, p<2e-16$; \llama{} 70B: $t=-57.193, p<2e-16$). The ordering of models in terms of predictive power generally aligns with standard notions of model quality: we find the weakest effect for \infinigram, and the effect sizes slightly increase with the parameter count of the neural models.

\Cref{fig:rating-vs-surprisal} shows the mean surprisal assigned to continuations in each condition on the x-axis, versus humans' subjective ratings on the y-axis. While the linear relationship captures the overall trend across conditions, it does not capture the relevant distinction between impossible and inconceivable: there appears to be substantial variation in LM surprisal for the Impossible and Inconceivable event continuations (yellow and red points), whereas humans rated most of these as having near-zero likelihood. We return to this issue in the Discussion (\Cref{sec:discussion}).

\section{Discussion}
\label{sec:discussion}

While a great deal of recent research in cognitive science has studied the distinctions between and within the improbable and the impossible, far less work has empirically or computationally examined people's reasoning about the inconceivable. Here, we investigated inconceivable events as a distinct mental category, as a way of studying more generally how people perform modal reasoning about (im)possibility. We found that people reliably distinguished inconceivable events from other modal categories, including impossible ones. 
These results expand on an ongoing research program in cognitive science and cognitive development that examines people's varying judgments of impossible events, and their distinction from the merely improbable \citep[e.g.,][]{komatsu_childrens_1986,browne_preschoolers_2004,shtulman_improbable_2007,shtulman_development_2009,lane_childrens_2016,goulding_similarity_2021,goulding_development_2023}. However, using a rating task, we also found that people did not distinguish impossible and inconceivable events through their subjective likelihood of occurring. We also found that neural language models -- but not a strong non-neural baseline -- distinguished between impossible and inconceivable event descriptions through their string probabilities. These probabilities broadly align with people's subjective ratings of event likelihood, but not within the categories of impossible and inconceivable events.

As mentioned in the Introduction, previous theoretical and empirical work points to two broad functional accounts of how impossibility and inconceivability might be distinguished: a distinction in \emph{kind}, or a distinction in \emph{degree}. For simplicity, we will refer to these two accounts as View 1 and View 2, respectively.
Given these two accounts, the novel finding that graded event probabilities (estimated by language models) can capture people's modal judgments could be explained in several ways: (a) people and LMs are performing convergent computations, either aligned with View 1 or View 2; or (b) people and LMs use different computations but arrive at the same outcome. We note that while our experiments do not fully tease these hypotheses apart, our findings are most consistent with a version of (b) where people do not make modal distinctions simply on the basis of event likelihood, but the statistics of language use are sufficient to separate descriptions of impossible and inconceivable events. We next discuss View 1 and View 2 in more depth, briefly surveying previous work that suggest \emph{a priori} reasons in favor of both accounts. After establishing this background, we then discuss potential explanations for our empirical findings.

Several lines of previous work in cognitive development \citep[e.g.,][]{shtulman_improbable_2007} and philosophy \citep[e.g.,][]{gendler_conceivability_2002, magidor_category_2013} align with View 1 (distinction in kind). Impossible events such as levitating a cow are nomologically impossible qua physics, but still conceivable as occurring in some possible world. By contrast, imagining something that is both a cow and not a cow is a logical violation that doesn't even get off the ground. Computationally, one proposal for a distinction in kind between impossible and inconceivable would be that judgments of inconceivability follow the experience of a category error, analogous to type errors in typed computer programs \citep{magidor_category_2009, sosa_type_2022}. To see this, consider the expression ``square\_root(45)''. Such an expression can be evaluated in different ways by different programs, depending on algorithm they use, but all reasonable programs designed to handle most math expressions will evaluate the expression. But, contrast the previous expression with ``square\_root(`dog')''. Such an operation will be rejected by many programs as a type violation, as `dog' is simply not the kind of thing you can apply the square-root program to. Some researchers have proposed that types, in the computer science sense that enforces the expected inputs and outputs of a program, may form the basis of mental computation across domains and behaviors \citep[e.g.,][]{morales2018representation,sosa_type_2022}. In much the same way that a computer program for calculating square-roots expects a number and would throw an error when encountering a string, the mind may expect ``baking a cake using'' to be followed by a `physical object' type, and throw an error if it is followed by the wrong type (e.g. ``the number three''). 
The argument would then go that inconceivable events are naturally categorized as such by the mind encountering a type error (``imagine levitating the number 5'' $\rightarrow$ type error $\rightarrow$ categorize as inconceivable), as opposed to impossible events, which \textit{can} be evaluated by a mental program. We note as an aside that given this view, and to the degree that the mind contains multiple modules equivalent to different programs, the same event or expression may be categorized as inconceivable or not, depending on the module that is evaluating it. 

This general view connects to a literature on type-theoretic distinctions and computations in semantic theory \citep[e.g.,][]{montague_proper_1973,luo_formal_2012,bekki_representing_2014,chatzikyriakidis_adjectival_2017,chatzikyriakidis_formal_2020,sutton_types_2024}.
Type theory and predicate logic provide building blocks which are fundamental to proposed compositions in natural language, such as objects, first- and higher-order predicates, properties, and propositions.
Building on this connection, a categorical distinction between impossibility and inconceivability has also been investigated from a language-processing perspective. In these studies, the general approach is to measure how people process sentences that violate different types of knowledge. Sentences that violate ``world knowledge'' correspond to impossible events, as we can evaluate them as being false in our world, while being able to conceive of possible worlds or circumstances in which they are true. However, sentences that violate ``word knowledge'' \citep[e.g., selectional restrictions;][]{chomsky_aspects_1965,katz_structure_1963} might correspond to inconceivable events, as it is difficult to derive a literal meaning for them by combining the lexical items.\footnote{We emphasize the difficulty of deriving \emph{literal} meanings. People can of course derive figurative or metaphorical meanings from sentences with apparent selectional restriction violations, and see also \citet{magidor_category_2013} for arguments that ``category error'' sentences do have meaning.} For example, consider the following sentences, adapted from \citet{warren_comprehending_2015}:

\ex. \a. The hamster lifted the refrigerator. \label{ex:hamster-lift}
    \b. The hamster entertained the refrigerator. \label{ex:hamster-entertain}

Given standard interpretations of the terms, the events in both \ref{ex:hamster-lift} and \ref{ex:hamster-entertain} are impossible: a hamster cannot \emph{lift} a refrigerator, nor can a hamster \emph{entertain} a refrigerator. However, \ref{ex:hamster-lift} is impossible because of our physical knowledge about the world, while \ref{ex:hamster-entertain} constitutes a selectional restriction violation (SRV) because refrigerators -- by virtue of being inanimate -- can't be entertained. Some theories of sentence comprehension propose that selectional restrictions are part of lexical knowledge, which is used more rapidly during the time-course of processing than world knowledge \citep[e.g.,][]{van_gompel_evidence_2005,bornkessel_extended_2006}. For example, eye-tracking studies have shown that SRVs appear to trigger processing difficulties that cannot entirely be explained by the plausibility or impossibility of the event \citep[e.g.,][]{rayner_effect_2004,warren_investigating_2007,warren_comprehending_2015}. From the neural perspective, violations of selectional restrictions have also been shown to elicit distinct patterns of event-related potentials, compared to general world knowledge \citep[e.g.,][]{paczynski_multiple_2012,sitnikova_two_2008}.\footnote{See \citet{mankowitz_category_2023} for arguments that category errors are neither necessary nor sufficient to trigger the N400 effect.} For example, \citet{paczynski_multiple_2012} find that world knowledge violations (e.g., ``The pianist played his music while the bass was strummed by the \underline{drummer} during the song'') evoke an N400 effect but no P600, whereas animacy-based SRVs (e.g., ``The pianist played his music while the bass was strummed by the \underline{drum} during the song'') evoke both an N400 and P600.
While our focus here is on modal reasoning rather than sentence comprehension, these studies provide behavioral and neural evidence for a distinction that is conceptually analogous to the distinction between impossibility and inconceivability.

In contrast to the separation in kind, View 2 suggests that the primary representation is a kind of hazy probability that never quite gets to zero, and has no principled distinctions of kind between impossible, improbable, and inconceivable. What we often term ``impossible'' is simply very improbable, and what we term ``inconceivable'' is simply very impossible. Further distinctions like ``violations of causal theories'' are then post-hoc rather than primary. If this view is true, it would be a rather radical departure from how modal distinctions have been proposed to operate in people. However, analogues of the graded view have been explored in the language processing literature. Prior studies have found that world knowledge violations elicit similar patterns of neural activity as SRVs \citep{hagoort_integration_2004,matsuki_event-based_2011}, and event-based plausibility plays a rapid and important role in online sentence comprehension \citep[e.g.,][]{garnsey_contributions_1997,mcrae_modeling_1998,kamide_time-course_2003,van_berkum_anticipating_2005}. Beyond this, the lexical/world knowledge distinction might also be dis-preferred on the basis of parsimony. Without theoretically motivated constraints, it seems there could essentially be as many selectional restrictions (or types, in the type-theoretic version of View 1) as there are verbs. For example, the verb ``to mail'' can only take objects that are ``mail-able'' \citep{myers_selectional_2005}, ``to inflate'' can only take objects that are ``inflate-able'', and so on \citep{mcrae_thematic_1997,matsuki_event-based_2011}. This becomes even more pronounced in our earlier example of ``levitating a feather with the number five'': ``levitating a feather'' can only be done using \emph{the kinds of things that can be used to levitate}, such as one's mind, a magical incantation, or an instantaneous gravity-reversing beam of energy.

Having discussed previous support for Views 1 and 2, we now turn to how they bear on our empirical results. Broadly, we found similarities in people's and LMs' behaviors: both people and LMs can distinguish modal categories, including impossible and inconceivable. One potential explanation is that people and LMs achieve this behavior through similar kinds of computations. Within this explanation, one sub-option is that they are both aligned with View 1: the primary computation is one of causal violations, type errors, intuitive theories, and other such distinctions of kind. The output of such computations then appears graded, due to probabilistic thresholds on top of these computations. The work surveyed above -- bridging semantic theory, cognitive science, and psycholinguistics -- suggests that, \emph{a priori}, these kinds of computations may support modal distinctions in people. If LM behaviors also operate this way, the internal computations and representations learned by language models in order to arrive in their judgments are quite sophisticated, having learned structured commonsense theories about domains such as biology and physics, and possibly type-based reasoning. If this view is true, it would be a rather radical departure from a common view in cognitive science that language models have not learned such representations, and it is also non-obvious how such models could be implementing type-based reasoning, dating back to the original debates surrounding connectionism and symbolic representations \citep[e.g.,][]{rumelhart_learning_1986,fodor_connectionism_1988,pinker_language_1988,ramsey_connectionist_1997}. While our findings do not definitively rule out that LMs are performing type-based reasoning, this view makes the least contact with our data, and would require substantial further investigation.

Another sub-option is that people and LMs both align with View 2, making distinctions between modal categories based on degree rather than kind. The threshold-probability view appears to be the best explanation for LM behaviors. Graded representations of probability are the currency of LM computation, and it is reasonable to expect LMs to implicitly learn generalized event knowledge as a by-product of the next-word prediction objective \citep[see, e.g.,][]{kauf_event_2023}. However, at first glance View 2 appears inconsistent with the findings of Experiment 2, where people did not distinguish between impossible and inconceivable event descriptions through their perceived likelihood of occurring. We note, however, it could be the case that people judge impossible and inconceivable events as having (near) zero chance of occurring, but they also represent graded notions of which inconceivable events are less likely than others. For example, perhaps if we directly asked people to rank inconceivable event descriptions in order of how likely they are, we would find structured agreement across people, just as prior work has found that people agree on some \emph{impossible} events being more impossible than others \citep{shtulman_explanatory_2017,mccoy_judgments_2019}. Such a finding would suggest that people use their commonsense knowledge to reason about events that are not only impossible, but cannot even be imagined in a possible world (without some process of non-literal interpretation or coercion). 

In contrast, the other high-level explanation is that people and language models are using different computations in a convergent fashion. Within this option it is also possible that people make modal distinctions based on graded representations, while LMs rely on distinctions of kind -- but this sub-option seems the least likely, given our findings in Experiment 2 (where people's likelihood ratings did not differentiate impossible and inconceivable events) as well as the current understanding of LMs, as discussed above. The other sub-option is that people make their modal distinctions in a manner of kind, consulting ``type errors'' and causal violations, even though event probabilities (as estimated through the statistics of language use) may in principle be sufficient for this distinction. This option seems the most consistent with our data. On the one hand, LMs are able to leverage the statistics of language use to assign probabilities that distinguish between impossible and inconceivable event descriptions. But on the other hand, people's likelihood judgments do not differ between impossible and inconceivable events, suggesting that people are using a different kind of computation to form this modal distinction. However, the caveats discussed above still hold -- for example, it could be that people would agree on the relative rankings within impossible or inconceivable events in a different task setting, and these rankings might correspond well to LMs' event description probabilities. We also have not provided direct evidence for type-based computations in humans, nor direct evidence \emph{against} type-based computations in LMs. It remains an open question how the type-error view could be cleanly separated from the threshold-probability view in an experimental setting.

In addition to comparing fully trained LMs to people's categorization and rating behaviors, we also found interesting patterns in LM training dynamics. Our investigation of modal categories across a single LM's training (summarized in Figure \ref{fig:surprisal-training}) suggests novel predictions about modal reasoning in children: the ability to distinguish between impossible and inconceivable should be present earlier than the ability to distinguish between improbable and impossible. This prediction could be tested by adapting the categorization experiments that we conducted with adults in Experiment 1, building upon prior categorization experiments in children \citep{shtulman_improbable_2007,shtulman_development_2009}. We stress that we are \textit{not} claiming direct parallels between model training and child development -- these processes substantially differ in many ways \citep{frank_bridging_2023}. Nevertheless, the trajectory of model training suggests high-level interactions between exposure to linguistic data and the emergence of key behavioral patterns \citep{zhang_when_2021,evanson_language_2023}. In particular, the earlier emergence of a distinction between the impossible and the inconceivable could inform ongoing broader debates about the development of modal reasoning: It could bridge between the emergence of basic modal reasoning and minimal representations of possibility \citep{leahy2020acquisition,leahy2022minimal, alderete2023three} and the later understanding of the distinction between improbable and the impossible. Further, to the degree that the later understanding rests on causal models or intuitive theories, the earlier emergence of the distinction would support the notion that basic modal possibility judgments may rely on other processes. 

Turning now to the AI perspective, our study contributes to a line of work seeking to understand the limitations and capabilities of language models -- in particular, what type of semantic, commonsense, or world knowledge LMs might learn through the objective of word prediction \citep[e.g.,][]{chang_language_2023,yildirim_task_2024}. The closest to our work is that of \citet{kauf_event_2023}, who found that LMs performed poorly at distinguishing between likely and unlikely events, whereas we found evidence that models robustly distinguished between probable and improbable events (\Cref{sec:exp2}). One potential reason for this disparity is that their likely/unlikely distinction involves social world knowledge, whereas our probable/improbable distinction relies on physical phenomena. For example, they tested materials such as ``The nanny tutored the boy'' (likely) versus ``The boy tutored the nanny'' (unlikely), or ``The actor won the award'' (likely) versus ``The actor won the battle'' (unlikely). It could be the case that the likelihood of events in these minimal pairs can be easily manipulated through context or are more similar \emph{a priori}, where the distinctions in our stimuli are sharper and less context-dependent. For example, there may be more contexts in which actors win \emph{battles} (e.g., against their competitors in a tough audition), but fewer contexts in which someone would chill a drink using \emph{snow}. Other studies have also shown that LMs perform poorly on ``counterfactual'' tasks which are possible but highly uncommon (conceptually similar to our Improbable condition), such as Python with 1-based indexing \citep{wu_reasoning_2024}. These results suggest that models' reasoning abilities are highly sensitive to specific patterns seen during pretraining \citep{mccoy_embers_2024}, instead of relying on more general task-solving skills. Another related study by \citet{sathe_language_2024} investigated whether LM conditional probabilities predict human acceptability judgments of novel, unattested collocations such as ``educated hairbrush''. However, they focused on a specific type of linguistic expression (adjective-noun pairings) and not on modal categories. To our knowledge, our experiments test LMs on fine-grained distinctions across the widest range of modal categories to date, and we demonstrate that LMs are able to reliably distinguish between the likely/unlikely, unlikely/impossible, and and impossible/inconceivable. These findings add to a growing body of evidence that structured knowledge about events and physical properties may be learned via statistical learning over linguistic forms. 

One potential limitation of our study is the validity of using LM-derived string probabilities to estimate the probability of an event. As discussed in the Introduction (\Cref{sec:intro}), this approach rests on an assumption that the likelihood of using a particular linguistic expression to describe an event is a proxy for how frequently the event occurs. This assumption faces at least two challenges, which we discuss in more detail here. First, let us suppose that language is used only to communicate about events that happen in the world. This view raises the issue of reporting bias \citep{sorower_inverting_2011,gordon_reporting_2013}: low-likelihood events may be over-represented in language, since people are incentivized to talk about them, whereas high-likelihood events may be underrepresented in language, since there is little communicative benefit of talking about them. Reporting bias is most likely to skew models' probability estimates of probable and improbable events, as events that do not occur in the world (i.e., impossible and inconceivable events) will not be described in language at all, if we adopt the view mentioned above. In our study, we do find that LMs assign lower probabilities (higher surprisals) to improbable than probable event descriptions, suggesting that modern LMs can overcome reporting biases to some extent \citep[cf.][]{shwartz_neural_2020}. 

The second challenge is related to non-literal uses of language. People use language not only to communicate about events literally happening in the world, but also to talk about fictional or hypothetical events, or to express figurative meanings. As a result, the probability an LM assigns to a sentence will be influenced by the non-literal language seen during training, in addition to factors that affect how likely the event is to literally occur. Therefore, there may be cases where a particular string completion makes an event impossible, but is still highly predictable given the context. For example, given ``The wizard plucked a scale from his fire-breathing'', the completion ``dragon'' might be extremely likely, even though dragons do not exist. This is not a major concern for our probable, improbable, and impossible stimuli, since the prefixes describe commonplace, physical events, which should set up expectations for physically plausible continuations. 

Non-literal language is, however, a potential concern for our inconceivable items, which are only inconceivable when interpreted in a strictly literal sense. For example, ``washing your hair with applause'' clearly violates physical knowledge, but could potentially be interpreted in a metaphorical manner by re-interpreting ``washing'' as a non-physical activity. We refer to these re-interpretations of inconceivable events as \emph{coercions}, mirroring the concept of type coercion in computer science. Indeed, a prominent theory of category mistakes argues that these types of sentences \emph{do} have meaning \citep{magidor_category_2013}, and \citet{chomsky_aspects_1965} writes that ``sentences that break selectional rules can often be interpreted metaphorically ... or allusively in one way
or another''. Another example of non-literal interpretation might apply to certain impossible event descriptions, such as ``baking a cake inside a freezer'', which could be interpreted as a phrase about cakes that are effectively ``baked'' inside refrigerators, such as tiramisu. This re-interpretation is potentially less metaphorical than it is about resolving communicative intent. For example, a listener might infer that a speaker chose to use the verb ``bake'' to convey the broad concept of preparing cakes, even though the literal definition of baking involves ovens or dry heat. While human participants in our experiments were told to consider the sentences in a strictly literal sense, it is not obvious how this restriction could be enforced when measuring string probabilities using language models, especially since models' training datasets are rife with non-literal language. 

Another limitation of our study is that we focus on one particular implementation of inconceivability: namely, the encountering of abstract concepts when a concrete noun is expected. But there is a broad range of ways in which events might be inconceivable: for example, by violating other features such as animacy (``the refrigerator giggled''), or by ascribing properties to objects in other odd ways (``the banana is an hour long'', ``the theory is blue''). It remains an open question whether these violations share some underlying cognitive process, or whether there are distinct avenues for dealing with different kinds of ontological oddities. In addition, we focused on judgments \emph{across} modal categories (probable, improbable, impossible, inconceivable), but it remains an open question how humans reason about items \emph{within} inconceivability. As discussed earlier, people make graded judgments among things that are nominally impossible, such as levitating a feather versus levitating a rock \citep[e.g.,][]{shtulman_explanatory_2017,mccoy_judgments_2019}. Similarly, there may also be structured gradedness within people's judgments of the likelihood of inconceivable items. Understanding the intuitive and commonsense knowledge that people use to make judgments about inconceivability is an important direction for future work.

Our focus in this work was on ``shades of zero'' -- distinctions within the region of extremely low probability, beyond people's or models' typical experience. While prior work has focused on distinguishing between the possible and impossible \citep[e.g.,][]{kauf_event_2023}, or the ability of language models to capture people's behavior on naturally occurring sentences \citep[e.g.,][]{smith_effect_2013,shain_large-scale_2023}, we believe that investigating the distinctions between impossibility and inconceivability is an exciting direction from both cognitive and AI perspectives, and can illuminate the boundaries of thought and imagination in both people and machines. In ancient maps, uncharted territories beyond the known world were marked with fantastical creatures that did not exist in the real world. There is much to explore in regions of extremely low probability. Here, there may literally be dragons. 

\section*{Acknowledgments}

This work has been made possible in part by a gift from the Chan Zuckerberg Initiative Foundation to establish the Kempner Institute for the Study of Natural and Artificial Intelligence. The experiments were supported by a grant from the Harvard University Hodgson Memorial Fund. We also acknowledge the support of the Jacobs Foundation (TDU). 

\bibliography{references}

\newpage

\appendix

\section{Additional analyses for Experiment 3}

\subsection{Item-level comparisons} \label{sec:appendix-exp3-item-level}

\begin{figure}[ht]
    \centering
    \includegraphics[width=\linewidth]{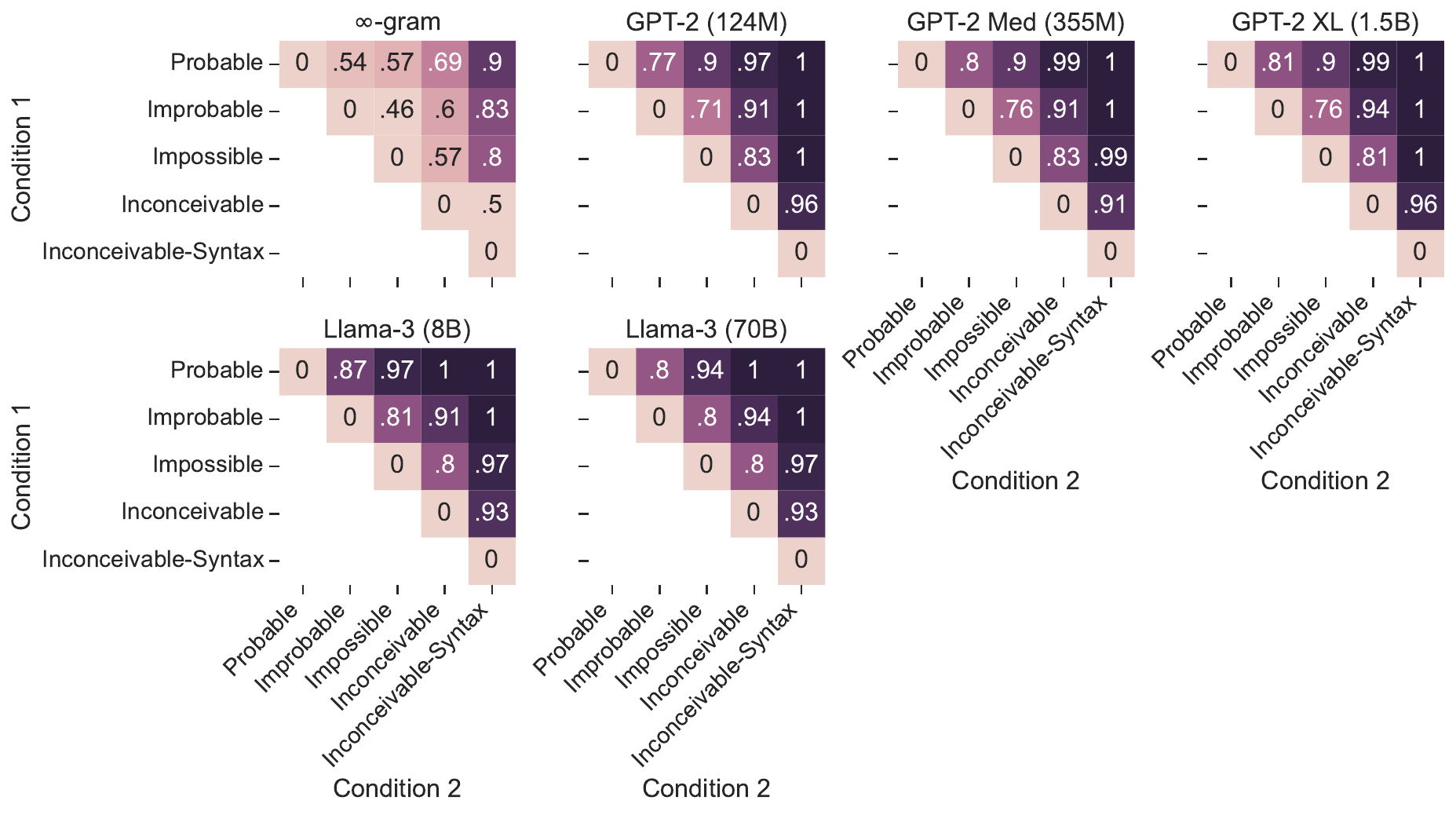}
    \caption{Item-level surprisal comparisons between pairs of conditions. Each cell (row $i$, column $j$) shows the proportion of items where the surprisal of the continuation in condition $i$ $<$ the surprisal of the continuation in condition $j$.}
    \label{fig:item-level-surprisal-heatmap}
\end{figure}

\Cref{fig:item-level-surprisal-heatmap} shows pairwise comparisons of surprisals across conditions within items. Each cell shows the proportion of items where the surprisal of the continuation in condition 1 (row) is less than the surprisal of the continuation in condition 2 (column).

\subsection{Full regression outputs} \label{sec:appendix-exp3-stats}

\Cref{tab:lmer} shows full results from the linear mixed-effects regression model fit for each language model in Experiment 2 (\Cref{sec:exp2}; \Cref{eq:lmer_condition_ngram}).

\begin{table}[ht]
    \centering
    \footnotesize
    \begin{tabular}{llrrrrrr} \toprule
Model & Predictor & Estimate & Std Error & DF & $t$ & $p$ & Sig. Code \\ \midrule
\infinigram & (Intercept)  & 5.59685  & 0.58427& 341.56437  & 9.579 & < 2e-16 & *** \\
\infinigram & Improbable > Probable  & 0.38345   & 0.19752& 275.68031  & 1.941 &  0.0532 & .   \\
\infinigram & Impossible > Improbable  & 0.10254   & 0.19723& 275.32756  & 0.520 &  0.6035 &  \\
\infinigram & Inconceivable > Impossible & 0.39321   & 0.19996& 278.51520  & 1.966 &  0.0502 & .  \\
\infinigram & Inconceivable-Syntax > Inconceivable  & 0.50355   & 0.25213& 316.95485  & 1.997 &  0.0467 & * \\
\infinigram & logNgramCount    & 0.19695   & 0.03541& 343.58475  & 5.561 & 5.39e-08 & *** \\ \midrule

\gpt & (Intercept)   & 1.85632    & 0.50714 & 321.16938  &  3.660 & 0.000294 & *** \\
\gpt & Improbable > Probable    & 0.88423    & 0.18125 & 274.68488  &  4.878 & 1.81e-06 & *** \\
\gpt & Impossible > Improbable    & 0.87134    & 0.18101 & 274.24529  &  4.814 & 2.45e-06 & *** \\
\gpt & Inconceivable > Impossible    & 1.50987    & 0.18327 & 278.22518  &  8.238 & 6.90e-15 & *** \\
\gpt & Inconceivable-Syntax > Inconceivable    & 1.99502    & 0.22689 & 325.82304  &  8.793 &  < 2e-16 & *** \\
\gpt & logNgramCount  & 0.22488    & 0.03084 & 326.24721  &  7.291 & 2.34e-12 & *** \\ \midrule

\gpt{} Med & (Intercept)  & 1.66233  & 0.49594 & 290.80265  & 3.352 & 0.000909 & *** \\
\gpt{} Med & Improbable > Probable   & 0.91011  & 0.18680 & 275.75602  & 4.872 & 1.87e-06 & *** \\
\gpt{} Med & Impossible > Improbable   & 0.98405  & 0.18658 & 275.27062  & 5.274 & 2.70e-07 & *** \\
\gpt{} Med & Inconceivable > Impossible   & 1.51907  & 0.18869 & 279.67165  & 8.051 & 2.38e-14 & *** \\
\gpt{} Med & Inconceivable-Syntax > Inconceivable   & 1.48940  & 0.22972 & 331.73046  & 6.484 & 3.25e-10 & *** \\
\gpt{} Med & logNgramCount    & 0.23270  & 0.03022 & 295.85151  & 7.701 & 2.04e-13 & *** \\ \midrule

\gpt{} XL & (Intercept)  & 1.45052  & 0.49455 & 344.00000 & 2.933 &  0.00358 & **  \\
\gpt{} XL & Improbable > Probable   & 0.86584  & 0.18872 & 344.00000 & 4.588 & 6.28e-06 & *** \\
\gpt{} XL & Impossible > Improbable   & 1.23210  & 0.18849 & 344.00000 & 6.537 & 2.27e-10 & *** \\
\gpt{} XL & Inconceivable > Impossible   & 1.50969  & 0.19057 & 344.00000 & 7.922 & 3.27e-14 & *** \\
\gpt{} XL & Inconceivable-Syntax > Inconceivable   & 2.02452  & 0.23108 & 344.00000 & 8.761 &  < 2e-16 & *** \\
\gpt{} XL & logNgramCount    & 0.24662  & 0.03014 & 344.00000 & 8.182 & 5.46e-15 & *** \\ \midrule

\llama{} 8B & (Intercept)  & 1.52189 & 0.51228 & 305.01804 & 2.971 &  0.00321 & **  \\
\llama{} 8B & Improbable > Probable   & 1.18717 & 0.18857 & 275.69218 & 6.295 & 1.20e-09 & *** \\
\llama{} 8B & Impossible > Improbable   & 1.30949 & 0.18833 & 275.22579 & 6.953 & 2.60e-11 & *** \\
\llama{} 8B & Inconceivable > Impossible   & 1.15916 & 0.19056 & 279.45135 & 6.083 & 3.87e-09 & *** \\
\llama{} 8B & Inconceivable-Syntax > Inconceivable   & 1.73243 & 0.23368 & 329.59096 & 7.414 & 1.05e-12 & *** \\
\llama{} 8B & logNgramCount    & 0.23494 & 0.03119 & 310.23786 & 7.533 & 5.48e-13 & *** \\ \midrule

\llama{} 70B & (Intercept) &  2.07629 & 0.53851 & 300.73340 & 3.856 & 0.000141 & *** \\
\llama{} 70B & Improbable > Probable  &  1.16023 & 0.19970 & 276.02326 & 5.810 & 1.72e-08 & *** \\
\llama{} 70B & Impossible > Improbable  &  1.40419 & 0.19945 & 275.55187 & 7.040 & 1.53e-11 & *** \\
\llama{} 70B & Inconceivable > Impossible  &  1.29915 & 0.20178 & 279.82344 & 6.439 & 5.23e-10 & *** \\
\llama{} 70B & Inconceivable-Syntax > Inconceivable  &  1.94857 & 0.24685 & 330.40583 & 7.894 & 4.37e-14 & *** \\
\llama{} 70B & logNgramCount   &  0.20858 & 0.03279 & 305.89998 & 6.360 & 7.32e-10 & *** \\ \bottomrule
\end{tabular}
    \caption{Results of linear mixed-effects regression model predicting surprisal values with fixed-effect of condition and log $n$-gram count, and random intercept of item (\Cref{eq:lmer_condition_ngram}).}
    \label{tab:lmer}
\end{table}

\end{document}